\documentclass[runningheads]{llncs}
\pdfoutput=1
\usepackage{amsmath}
\allowdisplaybreaks[4]
\usepackage{amsfonts}
\usepackage{amssymb}

\usepackage{graphicx}
\usepackage{subfigure}
\usepackage{algorithm}
\usepackage{algorithmicx}
\usepackage{algpseudocode}
\usepackage{multirow}
\title{\LARGE \bf
Intelligent Solution System towards Parts Logistics Optimization}
\author{Yaoting Huang \inst{1} \and Boyu Chen \inst{1} \and Wenlian Lu \inst{1} \and Zhong-Xiao Jin \inst{2} \and Ren Zheng \inst{2}}
\institute{Fudan University \and
SAIC Motor Corporation Limited}

\begin{document}

\maketitle
\thispagestyle{empty}
\pagestyle{empty}

\begin{abstract}
    Due to the complication of the presented problem, intelligent algorithms show great power to solve the parts logistics optimization problem related to the vehicle routing problem (VRP). However, most of the existing research to VRP are incomprehensive and failed to solve a real-work parts logistics problem.
    In this work, towards SAIC logistics problem, we propose a systematic solution to this 2-Dimensional Loading Capacitated Multi-Depot Heterogeneous VRP with Time Windows by integrating diverse types of intelligent algorithms, including, a heuristic algorithm to initialize feasible logistics planning schemes by imitating manual planning, the core Tabu Search algorithm for global optimization, accelerated by a novel bundle technique, heuristically algorithms for routing, packing and queuing associated, and a heuristic post-optimization process to promote the optimal solution.
	Based on these algorithms, the SAIC Motor has successfully established an intelligent management system to give a systematic solution for the parts logistics planning, superior than manual planning in its performance, customizability and expandability.

\end{abstract}

\section{Introduction}

Parts logistics optimization, aiming to solving the minimum cost of transporting all the required parts to its destination, is vital to modern industrial activities for all manufacturing enterprises, which is the key link of the supply chain management (SCM) and has been long studied \cite{VanderVaart2008}. In SCM researches, Supply chain integration is considered as a key factor of achieving improvement \cite{Tan1999,Romano2003} and have already become a powerful tool in real-world economic activities \cite{Olhager2004,Akintoye2000}.

The main component of parts logistics problem is the vehicle routing problem (VRP), introduced by\cite{dantzigTruckDispatchingProblem1959,anbuudayasankarSurveyMethodologiesTSP2014}, that is, designing optimal delivery collection routes from depots to a number of cities or customers, subjecting to side constraints \cite{Laporte1991}.
\cite{angelelliPeriodicVehicleRouting2002} researched multiple depots which vehicle can choose starting from.
\cite{crevierMultidepotVehicleRouting2007} introduced replenishment concept for VRPs, where vehicle can replenish capacity in several stations.
\cite{cordeauVRPTimeWindows1999} surveyed VRPs with time window-constraints (VRPTW). 
\cite{fleischmannVehicleRoutingProblem} considered multiple use of vehicles (VRPMU).
\cite{ioriExactApproachVehicle2007} appended 2-dimensional loading constraints to VRP (2L-CVRP) and use branch-and-bound for checking loading feasibility.
\cite{gendreauTabuSearchHeuristic2008} utilized tabu search to solve 2L-CVRP. However, above researches considered constraints separately. Due to this circumstances SAIC motor is still utilizing old-fashion manual planning scheme.

This work incorporates above works and hereby proposes the 2-Dimensional Loading Capacitated Multi-Depot Heterogeneous Vehicle Routing Problem with Time Windows.

\section{Problem Formulation}
\subsection{Problem Scenario Description}
\begin{figure}
	\centering
	\includegraphics[scale=0.22]{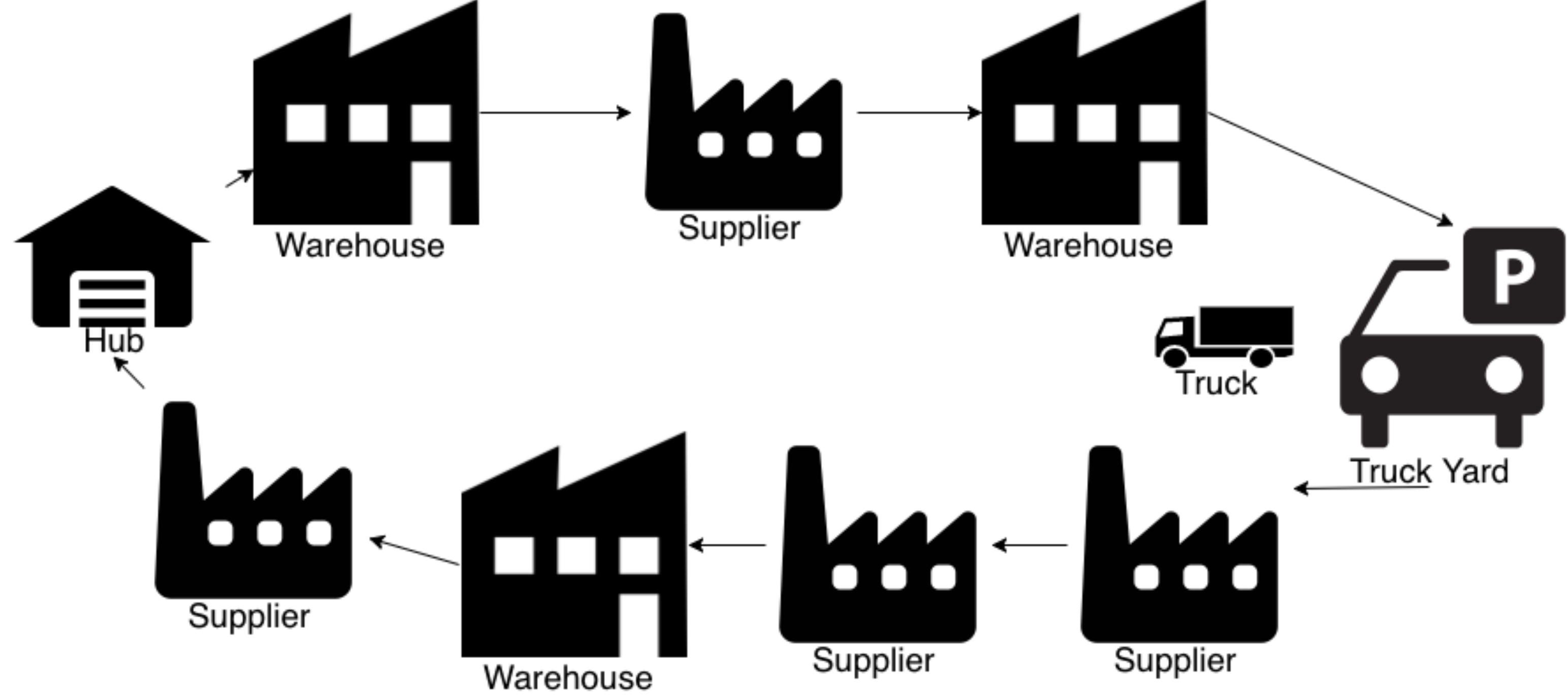}
	\caption{An example route: a truck departure from a vehicle yard, visiting suppliers with loading shipments, warehouse and the hub with unloading shipments according to the planning in an intermittent way, and finally ending its daily task by returning to the truck yard}
	\label{fig:ilu}
\end{figure}
Typical parts logistics are briefly described as follows: the supply chain system requests parts from various suppliers to be delivered to plant warehouses for assembly. Herein, the basic transportation unit is named as {\em shipment}. Each shipment is composed of a bunch of same type bins, which should be picked up from a specific supplier and be delivered to a specific plant warehouse. These transportation is carried out by the logistic department with adequate trucks of different models, starting off and ending from a specific vehicle yard. The pickup and load at the plant warehouse occur at its limit number of docks, each of which allows several specific models of truck to pickup/load at specific given time-windows. These activities of loading and unloading cost fixed time lengths to be completed. A hub is also considered in this problem, which is able to receive scattered shipments to be integrated. Figure.\ref{fig:ilu} illustrates a typical route by a truck.

The basic unit of loading/unloading is bin. The 3D packing problem is simplified to a 2D packing problem, As the bins should initially be stacked into columns following stacking rule or with pallets. The information of a shipment contains: supplier, plant warehouse, bin quantity, bin size (length, width and height), stacking layer limitation, pallet requirement, pickup time interval and delivery time-interval.

The complication of this problem owes to the numerous constraints. The main constraints are of the {\em Time-window constraint} (TW). 
\begin{itemize}
\item[A1.] 
Suppliers and plants have working time-windows so that the whole time-interval of the visit of the truck, including the loading or unloading should be contained in this working time-window. 
\item[A2.] According to the limit of the dock of the supplier and warehouse, there should be only limited number of trucks visiting this specific depot at the same time. More trucks than this number leads queueing.
\item[A3.] Each shipment has its pickup and delivery time-intervals that should be satisfied.

\end{itemize}
Another important constraints are of {\em loading/unloading constraints}.
\begin{itemize}
\item[B1.] Bin stacking should follow given rules: the bins of the same type from the same supplier can be stacked together and some specific types must be loaded on pallets, which can be stacked. The maximum number of layers of bins/pallet is limited.
\item[B2.] The bins must be contained within the loading surface of the truck, and no two bins can overlap.
\item[B3.] Because the shipments will be handled by forklift, sequence constraint of loading should be considered\cite{Iori2007}: when a location is visited, and the bins of the corresponding lot, can be downloaded or uploaded through a sequence of straight movement parallel to the width of the loading area.
\end{itemize}
There are additional essential constraints : according to the company's rule, the shipments from different cities cannot be loaded on a single truck; Docks of some suppliers restrict the truck lengths. Some suppliers restrict the number of visit times. A very few suppliers request to be the first to be visited site of on a route and some a few warehouses request to be the last. We add judgement statement in coding and won't discuss them further.

In overview, the followings are decisions to decide a feasible planning:
\begin{itemize}
	\item Select which warehouses and suppliers a truck visits;
	\item Plan the route for this truck;
	\item Pack specific bins with constraints.
\end{itemize}
We utilize total mileage as cost function, and our object is to minimize it.

\subsection{Mathematical Model}
\begin{table}
	\scriptsize
	\renewcommand\arraystretch{1.1}
	\caption{The parameters and decision variables}
	\label{table:var}
	\centering
	\begin{tabular}{lp{.7\textwidth}}
		\hline
		\bfseries Parameter/ & \bfseries Definition\\
		\bfseries decision variable& \\
		\hline
		$C_i$ &  Cost per mileage unit of truck $i$\\
		$t_{ik}$ & Departure time of truck $i$ at $k$th location\\
		$w_{ik}$ & Waiting time of truck $i$ at $k$th location\\
		$n_i$ & Number of locations truck $i$ should visit\\
		$N_S$ & Number of shipments\\
		$d_j$ & Location order of a route the  shipment $j$ will be delivered\\
		$p_j$ & Location order of a route the  shipment $j$ will be picked up\\
		$P_j$ & If shipment $i$ need pallet then $P_j=1$, otherwise 0\\
		$u_{j\theta}, v_{j\theta}$ & Width and length coordinate of the $\theta$th column of shipment $i$ on a truck\\
		$u_{j\theta\gamma}, v_{j\theta\gamma}$ & Width and length coordinate of the $\gamma$th column on $\theta$th pallet, only valid when $P_j=1$\\
		$W_j, L_j$ & Bin width and length of shipment $i$\\
		$W^P, L^P$ & Pallet width and length\\
		$W^V_i, L^V_i$ & Width and length of truck $i$\\
		$n^s_j$ & Number of column stacked by shipment $j$\\
		$b^p_{j\theta}$ & Number of column in pallet $\theta$, only valid when $P_j=1$\\
		$N^B_j$ & Number of bins of shipment $j$ \\
		$L_j$ & Bin stack layer limitation of shipment $j$ \\
		$h_{jj'}$ & Equals 1 if shipment $j,j'$ was the divided shipment, and destination of $j$,  source of $j'$ are the hub, otherwise 0\\
		$D(\zeta , \eta)$& Distance between two locations $\zeta$ and $ \eta$\\
		$TW(\iota)$ & Working time interval of a location $\iota$\\
		$T(\zeta, \eta)$ & Time cost between two locations $\zeta$ and $ \eta$ \\
		$TH(\iota)$ & Handling time in a location  $\iota$ \\
		$\Psi(n, A)$ & The smallest of the $n$ largest  number in set $A$\\
		$DC(\iota)$ & Dock number of a location $\iota$ \\
		$TP(\omega)$ & Pickup time interval of a shipment  $\omega$\\
		$TD(\omega)$& Delivery time interval of a shipment  $\omega$ \\
		\hline
	\end{tabular}
\end{table}
We define $X = (x_{ij})$ as the truck-shipment relations, where $x_{ij} = 1$ indicate that shipment $j$ will be handled by truck $i$, otherwise $x_{ij} = 0$,  and we define $y_{ik}$ as the $k$th station the truck $i$ will visit. Other symbols of the mathematical model are defined in Table \ref{table:var}. Constraints are listed in TABLE \ref{table:con}. Thus, following the description above, the parts logistics planning is formulated by the following optimization problem:
 \begin{equation}
 	\text{Minimize} \quad \sum_{i=1}^{N}\sum_{k=0}^{n_i} C_i D(y_{ik}, y_{i(k+1)})
 \end{equation}

\section{Solution}
\begin{figure*}
	\centering
	\includegraphics[scale=0.40]{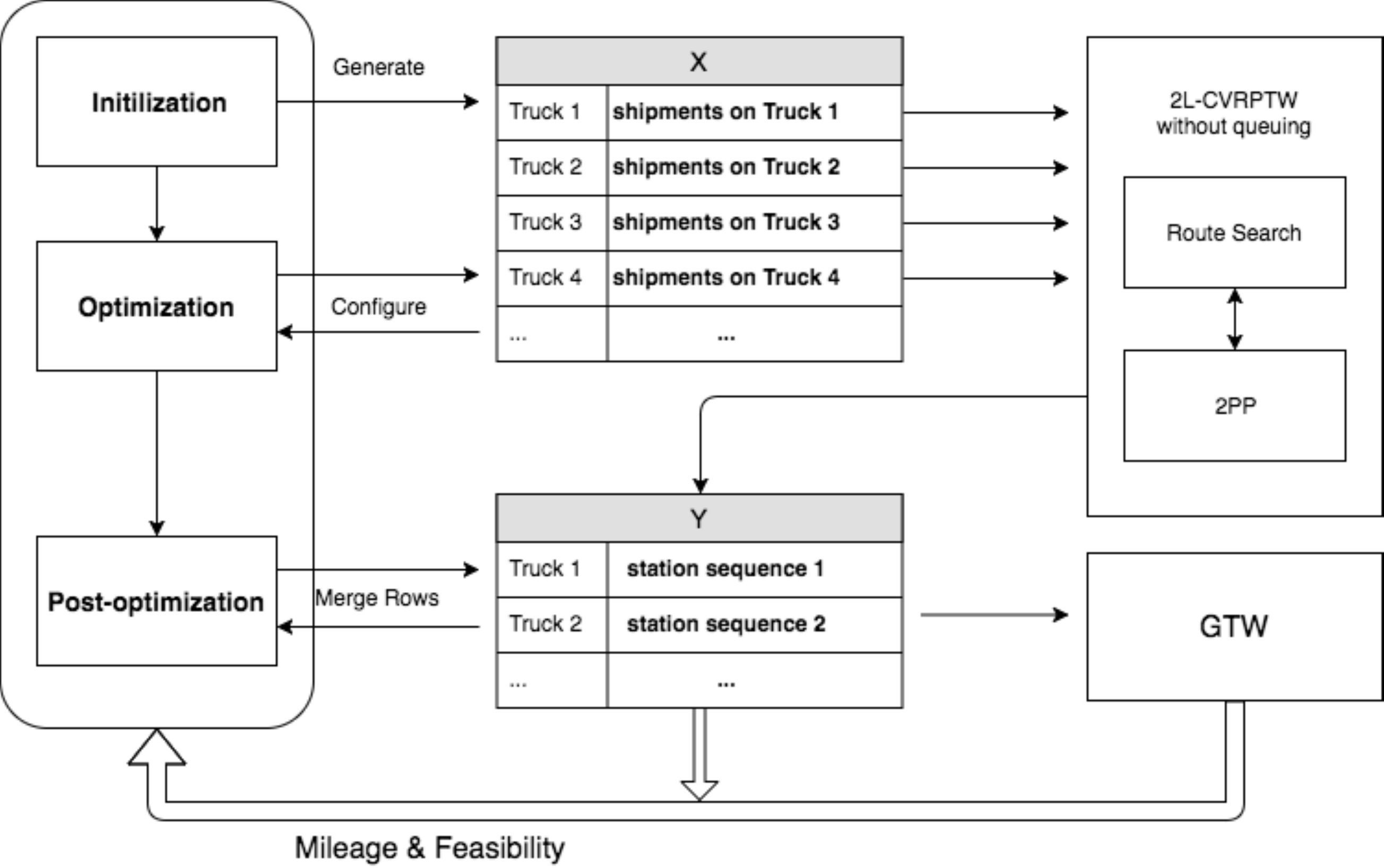}
	\caption{Overview of Algorithm Structure, Initialization and optimization will only configure shipment-truck relations $X$, which will then generate routes with A1, A3 and B series constraints, after that global time window (GTW) feasibility (A2 constraint) will be checked. The post-optimization process will configure $Y$ for further improvement}
	\label{fig:structure}
\end{figure*}
\subsection{Algorithm Architecture}

We divide the whole algorithm into three procedures: initialization, optimization and post-optimization. Figure.\ref{fig:structure} illustrates the work-flow underlying algorithms: the Initialization is to generate a feasible solution; The optimization, which in our case is tabu search, adopts this initial solution and optimize at the shipment level towards minimizing the total mileage. The post-optimization further improves at the route level. In addition, there are an axillary {\em 2-Dimensional Loading Capacitated VRP with Time Window (2L-VRPTW) solver}, which generates a route of the given truck and shipments and the associated packing schemes.

\subsection{2L-VRPTW Solver}
Given a $X$, 2L-VRPTW Solver generate route and packing scheme for each truck satisfying A1, A3 and B series constraints. The algorithm goes as following: a asynchronous route search procedure search all the routes space, which is built as a tree. Branches violating A1 and A3, or its partial mileage is larger than searched feasible leaf node will  be pruned.

When we reach the leaf node (representing a complete route), we will judge feasibility with respect to B series constraints. This packing procedure is conducted by the following steps: first, stacking the bins and pallets into columns; second, pre-justifying whether the packing at this site is possible by preassigned threshold of the ratio of the coverage area of the columns over the total; If the pre-justification is passed, 2-D Packing Problem (2PP) is solved by searching following a heuristic function with respect to the wasted area, convexity and covered area. We reduce the search space by limiting the search width.

\subsection{Initialization}
A good initialization scheme is generally believed to be essential to the optimization algorithm. Herein, we initialize by imitating the manual planning: First, we cluster all the suppliers into several disjoint areas by a community detection algorithm \cite{Martelot2013}, and then subgroup all shipments into several subsets according to this supplier area clustering. Within a shipment subset, we sort the shipments in a descending order based on the distance between supplier and warehouse. Then we assign trucks to load these shipments in a one-by-one manner unless some constraints are not satisfied.

Towards agreement with the A2 constraint, when finding truck violating, we delete the existing truck, and add all its assigned shipments into a sequence. After all other shipments are assigned, we deal with the sequence with the same algorithm. This process will be done several times until all truck satisfying A2.

\subsection{Optimization --- Tabu Search}

We implement tabu search (TS) algorithm for optimization. TS can be illustrated as choosing the optimal neighborhood solution and declare the reverse move tabu. However, TS need fully evaluation of all neighbors at each step, thus we need to carefully define the neighbor so that the computation complex is not too large and the algorithm can converge faster.

The neighborhood is given as follows: a new solution $X'$ is generated from given $X$ by moving a shipment from one truck to another truck. This operation is costly with numerous shipments. To reduce the complexity, the shipments by the same supplier and warehouse destination are bundled, and the movement is restricted within the bundle. We denote the bundle movement, from $X$ to $X'$, $(X \rightarrow X')$.
\begin{figure}[h]
    \centering
    \includegraphics[scale=0.15]{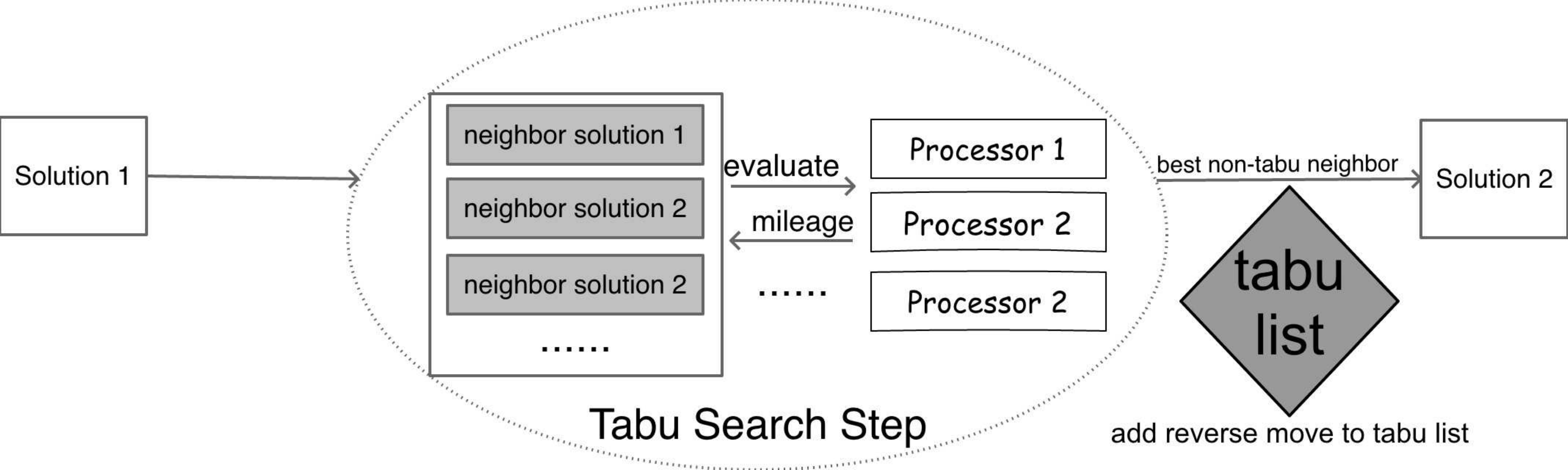}
    \caption{Tabu Search}
    \label{fig:ts}
\end{figure}

The optimization process fetches initial solution from initialization module, then bundles all the shipments. At each TS step, all neighborhood solutions will be evaluated asynchronously, and the reverse bundle move will be declared tabu. We also keep track of the best solution so far. The overall algorithm is shown in Following algorithms.

\begin{algorithm}
	\scriptsize
	\caption{TS Optimization}
	\begin{algorithmic}[1]
		\Require Initial scheme $X_0$, bundle break threshold $B$
		\State Bundle all shipments
        \State initial tabu list $\mathcal{T}$
		\State set $X=X_0$, best solution keeper $X^*$
        \Repeat
		\State Evaluate all non-tabu neighbor solution $\{X_k\}$ of $X$, where $(X \rightarrow X_k) \notin \mathcal{T}$
		\State Choose $X' \in \{X_k\}$ with the least mileage
		\If{mileage of $X' \leq$ mileage that of $X^*$}
		\State $X^* \leftarrow X'$
        \EndIf
        \State $\mathcal{T} = \mathcal{T} \cup \{(X \leftarrow X')\}$
		\State $X \leftarrow X'$
		\Until{given computation resorces reached}
		\State \Return $X^*$
	\end{algorithmic}
	\label{alg:ts}
\end{algorithm}

\subsection{Post-optimization}
Post-optimization process aims to optimizations which are hard to consider in main optimization process and brings little descent of cost. Firstly, we will try to replace larger truck with smaller one. Then we try to merge routes where a truck will run the sub-routes one after another, so it will not return to its depot after one sub-route is done, with the cost of occupying more time windows. It is done by heuristic algorithm, where we do emergence until there are no mergeable route.
To determined the sub-routes sequence. Here we define distance from one sub-route to another as the distance from one sub-route's non-yard destination to another one's origin. Thus, the problem can be viewed as a common Travel Salesman problem (TSP), and we solve it by dynamic programming \cite{Bellman1962}.

\section{Experiment and Result}
We tested on a data set of 311 shipments obtained from SAIC logistic division, containing 54 variants of bins, alone with the data of 45 suppliers and 8 plant warehouses. We run the test on Intel Core i7 CPU \@ 2.70GHz*4. We ran TS for 5000s and compared TS with (TS-WB) and without (TS-NB) shipment bundle. Table.\ref{tab:tab} shows the results.
\begin{table}[h]
	\centering
	\scriptsize
	\caption{Post-optimization results}
	\label{tab:tab}
	\begin{tabular}{|c|c|c|c|}
		\hline
		                     & Original & After Post-opt. & Diff.  \\
		\hline
        Initialization     & 1131.80  & 1099.40               & -32.40 \\
        \hline
        TS-WB     & 503.30  & 482.50                & -20.80   \\
        \hline
		TS-NB & 1014.60  & 1014.60                & 0.00 \\
		\hline
	\end{tabular}
\end{table}

\begin{figure*}[h]
	\centering
	\subfigure[Bundle Efficiency]{\includegraphics[scale=0.29]{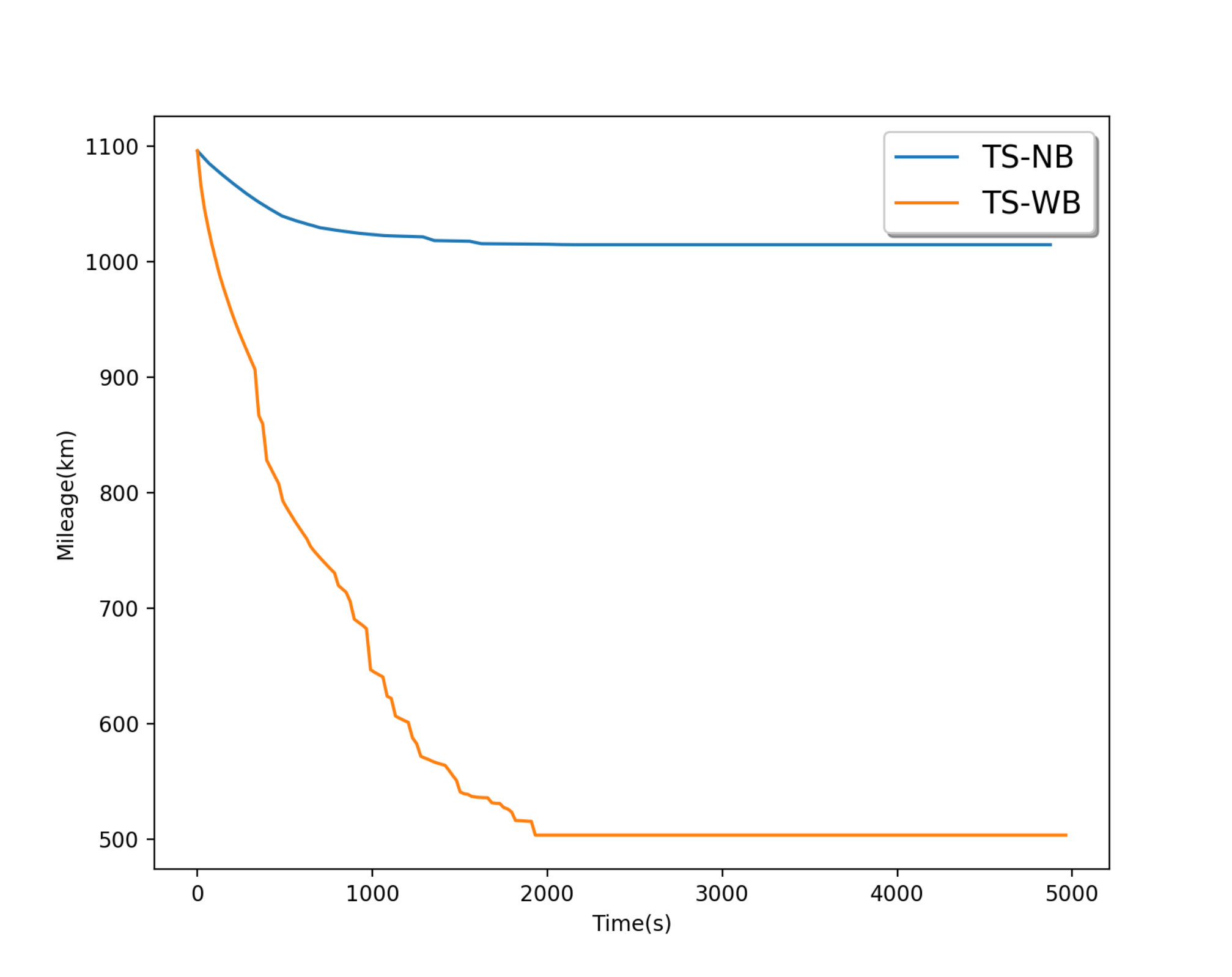}}
	\subfigure[Robustness]{\includegraphics[scale=0.375]{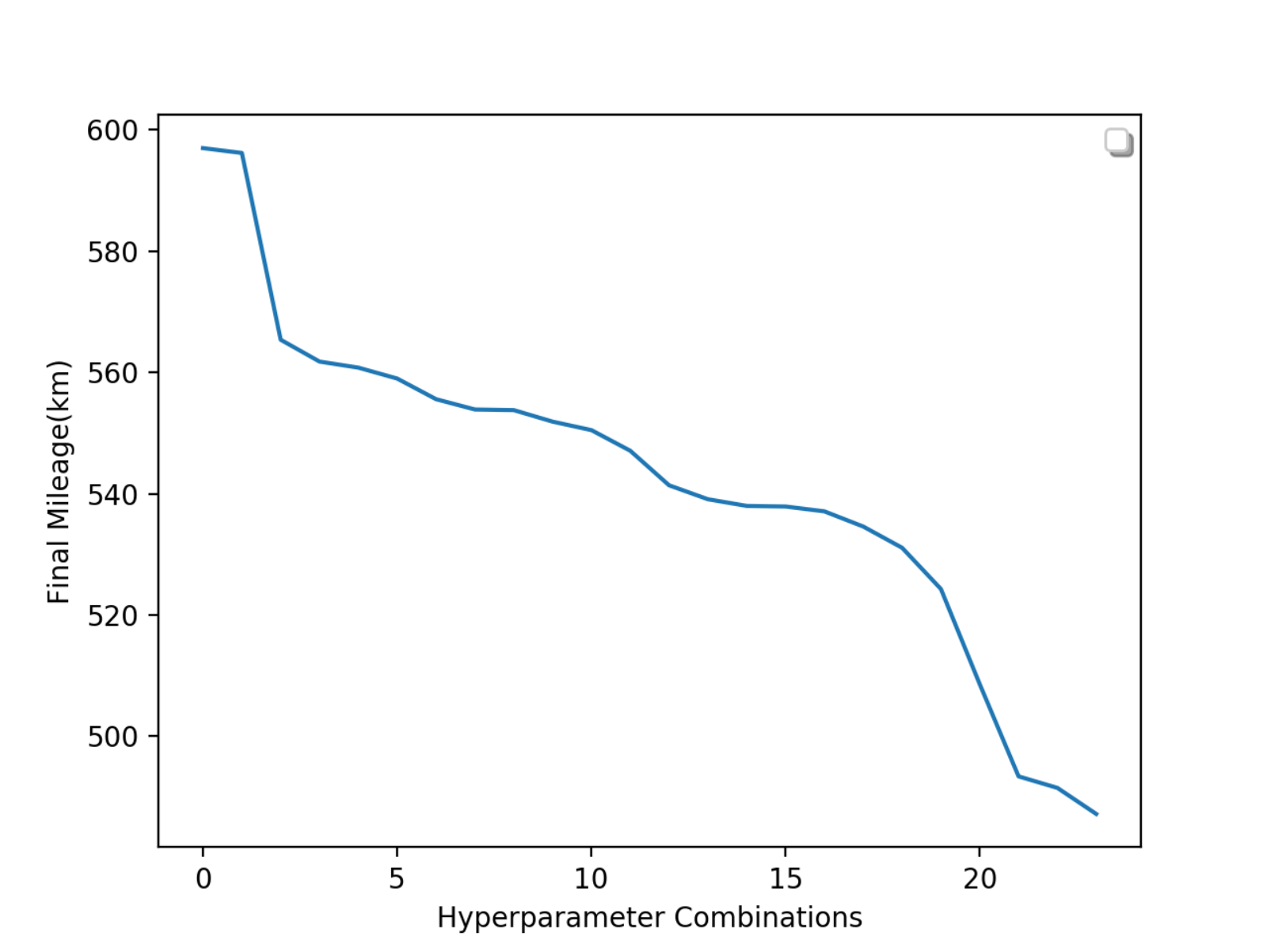}}
	\caption{Results of optimization process. (a) shows the mileage descend process with bundle and without bundle.(b) shows the robustness across 24 hyperparameter combinations (all with bundle). x-axis indicated ranking of hyperparameter combination in descent order, and y-axis indicate the total mileage of each combination}
	\label{fig:converge}
\end{figure*}

In Table.\ref{tab:tab}, we can see that post-optimization can only have limited effect. This is because optimization and pot-optimization compete for time-window occupation. However, Post-optimization is necessary, because in circumstance when a massive amount of shipments required between two very close stations, it is always better to arrange one truck carry these shipments back and forth without going back to truck yard, such circumstances is difficult to be considered in optimization process.

The mileage changing curve with respect to time is shown in Figure.\ref{fig:converge} (a). We can see that the bundle technique, leading to larger TS step, made it easier to escape local minima, ended in better convergence. Robustness experiment in Figure.\ref{fig:converge} (b) shows that, TS algorithm is very robust, referring to initial mileage scale.

\section{SAIC Logistic Management System (SPRUCE)}
Our research is assisted by SAIC motor, and adopted into their auto part logistic management system. The Shanghai Auto Incorporate Company (SAIC) Motor Co. Limited is the largest auto maker of China and has its own supply chain system that contains more than $500$ suppliers, $4$ plants and delivers more than $10000$ shipments daily. 

The algorithm above is utilized for the SAIC Motor to build up a parts logistics scheme management system consisting of several modules. Shipment Management Module accepts new shipment and mange them. Data Maintenance Module maintains station and truck states, pack parts into bins. Global Optimization Module generates and optimizes routing scheme. Graph Generator Module generates routing map and stowage plan for operational usage. Manual Planning Module handles temporary shipment manually.

\begin{figure}
	\centering
	\includegraphics[scale=0.33]{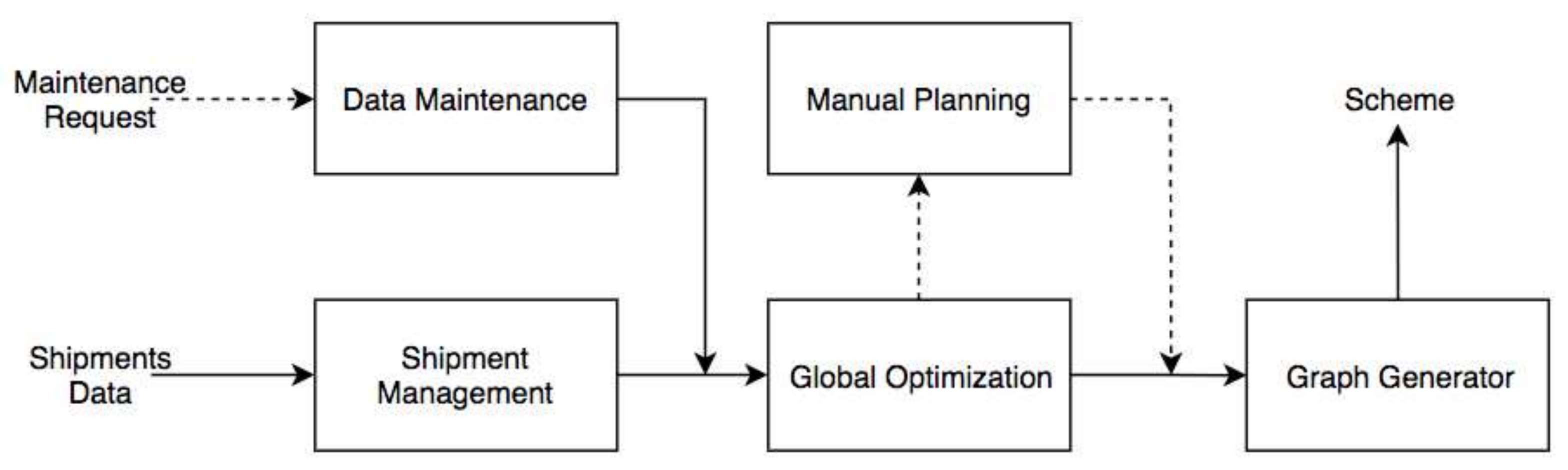}
	\caption{SAIC Logistic Management System}
	\label{fig:system}
\end{figure}

The system, illustrated in Figure \ref{fig:system}, accepts raw shipments data and parameter maintenance request, then transfers them into Global Optimization Module. This Module will possibly utilize Manual Planning Module to Handle temporary shipments and very important shipments. The results will be eventually processed by Graph Generator to routing map, stowage plan, time plan, and schedule of truck resources. The whole system can handle 2000 shipments in about 10min to 15min.

\section{Conclusion}
In this paper, we established a parts logistics optimization model, which is mathematically a 2-Dimensional Loading Capacitated Multi-Depot Heterogeneous Vehicle Routing Problem with Time Windows, and presented algorithms for its systemic solution, by using TS  accelerated by pruning methods and shipment bundling techniques. This systemic solution has shown efficient power to the optimization problem and has been utilized for the SAIC to establish its parts logistics scheme management systems

We will concentrate on further accelerating the computation. For instance, for the 2PP, a deep Q-learning method is being used and another effort is taken to incorporate genetic algorithm by utilizing parallel computing.

\begin{table*}
	\tiny
	\renewcommand\arraystretch{1.2}
	\caption{Constraints}
	\label{table:con}
	\centering
	\begin{tabular}{p{.3\textwidth}|p{.7\textwidth}}
		\hline
		\bfseries Constraint & \bfseries Formula\\
		\hline
		\multirow{5}*{Working TW (A1)} & $\text{for } \forall i \in \{1,2,\cdots, N\}, \forall k \in \{0,1,\cdots, n_i\}$:\\
		&$y_{i0} = y_{in_i} = y_0$ \\
		& $t_{i0} = 0$ \\
		& $t_{i(k+1)} = t_{ik} + T(y_{ik}, y_{i(k+1)}) + w_{i(k+1)}$ \\
		& $t_{ik} + T(y_ik, y_{i(k+1)}) \in TW(y_{i(k+1)})$ \\
		\hline
		\multirow{3}*{Queue and dock(A2)}& $\text{for } \forall i \in \{1,2,\cdots, N\}, \forall k \in \{0,1,\cdots, n_i\}$:\\
		& $w_{i(k+1)} = TH(y_{i(k+1)}) + (\Psi(DC(y_{i(k+1)}, \{t_{i'(k'+1)} | t_{i'k'} + T(y_{i'k'}, y_{i'(k'+1)}) \leq t_{ik} + T(y_{ik}, y_{i(k+1)})  \})) - (t_{ik} + T(y_{ik}, y_{i(k+1)})) )$ \\
		\hline
		\multirow{3}{5cm}{Shipment TW(A3)}& $\text{for } \forall j \in \{0,1,\cdots,N^S\}:$\\
		& $\sum_{i=1}^{N} t_{ip_j} x_{ij} \in TP(\sum_{i=1}^{N} y_{ip_j} x_{ij}),\sum_{i=1}^{N} t_{id_j} x_{ij} \in TD(\sum_{i=1}^{N} y_{id_j} x_{ij})$ \\
		&$p_j < d_j$\\
		\hline
		\multirow{10}{5cm}{Stacking Rule (B1)}& $\text{for } \forall j \in  \{0,1,\cdots,N^S\}, \forall \theta \in \{0,1,\cdots,n^s_i\}$:\\
	& $\sum_{\theta} l_{j\theta} = N^B_j, \forall j \in \{0,1,\cdots,N^S\}$\\
	& $l_{j\theta} \leq L_j$\\
	& $\text{ for } \forall j, \theta,  \text{ and } \gamma \in \{0,1,\cdots,b^p_\theta\} \text{ and } P_j = 1$:\\
	& $0 < u_{j\theta \gamma} \leq W^P - W_j , 0 < v_{j\theta \gamma} \leq L^P - L_j $\\
	&$NB_j = \sum_{\theta}\sum_{\gamma} l_{j \theta \gamma}$\\
	&$ l_{j\theta \gamma} \leq L_j$\\
 	& $\forall \gamma \neq \gamma'$one of the following must be true:\\
	&$u_{j \theta \gamma'} \geq u_{j\theta \gamma} + W_j, u_{j \theta \gamma}  \geq u_{j\theta \gamma'} + W_j$ \\
	&$ _{j \theta  \gamma'} \geq v_{j\theta  \gamma} + L_j, u_{j \theta  \gamma} \geq v_{i\theta  \gamma} + L_j $ \\
	\hline
	\multirow{8}{5cm}{2-D loading(B2)}&$\text{for } \forall j,j' \in  \{0,1,\cdots,N^S\}, \forall \theta, \theta' \in \{0,1,\cdots,n^s_i\}$:\\
	& $0 < u_{j\theta} \leq\sum_{i} (x_{i'j'} W^V_i x_{ij} - W_j) $ \\
	& $ 0 < v_{j\theta} \leq\sum_{i} (x_{i'j'} L^V_i x_{ij} - L_j) $ \\
	& one of the following must be true:\\
	& $\sum_{i} x_{ij'}u_{j' \theta'} \geq \sum_{i} x_{ij}(u_{j\theta} + W_j(1-P_j) + P_jW^P)$ \\
	& $\sum_{i} x_{ij}u_{j \theta} \geq \sum_{i} x_{i'j'}(u_{j'\theta'} + W_{j'}(1-P_j') + P_{j'}W^P)$ \\
	& $\sum_{i} x_{ij}v_{j' \theta'} \geq \sum_{i} x_{ij}(v_{j'\theta'} + L_j(1-P_j) + P_{j}L^P) $ \\
	& $\sum_{i} x_{ij}v_{j \theta} \geq \sum_{i} x_{i'j'}(v_{j'\theta'} + L_{j'}(1-P_j') + P_{j'}L^P) $ \\

	\hline
	\multirow{4}{5cm}{Loading sequence(B3)}& $\text{for all } j \neq j' \text{ and } d_j' < d_j \text{ and } p_j \leq p_j'$ one of the following must be true:\\
	& $\sum_{i} x_{ij}u_{j \theta} \geq \sum_{i} x_{i'j'}(u_{j'\theta'} + W_j'(1-P_j') + P_{j'}W^P) $ \\
	& $\sum_{i} x_{ij}v_{j' \theta'} \geq \sum_{i} x_{ij}(v_{j'\theta'} + L_j(1-P_j) 	+ P_{j}L^P) $ \\
	& $\sum_{i} x_{ij}v_{j \theta} \geq \sum_{i} x_{i'j'}(v_{j'\theta'} + L_j'(1-P_j') + P_{j'}L^P) $ \\
	
	\hline
	\multirow{1}{5cm}{Hub shipments}&$\text{for }\forall j\neq j':0 \leq [\sum_i t_{i p_j} x_{ij} - \sum_i t_{i p_j} x_{ij}] h_{j j'}$\\
	\hline
	\multirow{1}{5cm}{Shipments must be loaded}&$\text{for } \forall j \in \{0,1,\cdots,N^S\}: \sum_{i=1}^{N} x_{ij} = 1  $\\
	\hline
	\end{tabular}
\end{table*}
\end{document}